\documentclass{article}

\usepackage[accepted]{icml2021}

\usepackage[utf8]{inputenc} 
\usepackage[T1]{fontenc}    
\usepackage{hyperref}       
\usepackage{url}            
\usepackage{booktabs}       
\usepackage{amsfonts}       
\usepackage{nicefrac}       
\usepackage{microtype}      

\usepackage{natbib}
\usepackage{graphicx}
\usepackage{subfigure}
\usepackage{amssymb}
\usepackage{todonotes}
\usepackage{amsmath}
\usepackage{makecell}
\usepackage{tikz}
\usepackage[outline]{contour}
\usepackage{nccmath}
\usepackage{caption}     
\usepackage{multirow}
\usepackage{fix-cm}
\usepackage[algo2e]{algorithm2e}

\usepackage{amsmath}
\usepackage{graphicx}

\usepackage[normalem]{ulem}
\useunder{\uline}{\ul}{}

\begin{document}

\twocolumn[
\icmltitle{CRUDE: Calibrating Regression Uncertainty Distributions Empirically}
\icmlsetsymbol{equal}{*}

\begin{icmlauthorlist}
\icmlauthor{Eric Zelikman}{}
\icmlauthor{Christopher Healy}{}
\icmlauthor{Sharon Zhou}{}
\icmlauthor{Anand Avati}{}
\\Stanford University\\
\small{\texttt{\{ezelikman, cjhealy, sharonz, avati\}@cs.stanford.edu}}
\end{icmlauthorlist}


\vskip 0.3in
]

\printAffiliationsAndNotice{}  


\newcommand{\X}{\mathcal{X}}
\newcommand{\E}{\mathbb{E}}
\newcommand{\V}{\mathbb{V}}

\begin{abstract}
Calibrated uncertainty estimates in machine learning are crucial to many fields such as autonomous vehicles, medicine, and weather and climate forecasting. While there is extensive literature on uncertainty calibration for classification, the classification findings do not always translate to regression. As a result, modern models for predicting uncertainty in regression settings typically produce uncalibrated and overconfident estimates. To address these gaps, we present a calibration method for regression settings that does not assume a particular uncertainty distribution over the error: \textit{Calibrating Regression Uncertainty Distributions Empirically} (CRUDE). CRUDE makes the weaker assumption that error distributions have a constant arbitrary shape across the output space, shifted by predicted mean and scaled by predicted standard deviation. We detail a theoretical connection between CRUDE and conformal inference. Across an extensive set of regression tasks, CRUDE demonstrates consistently sharper, better calibrated, and more accurate uncertainty estimates than state-of-the-art techniques.
\end{abstract}

\section{Introduction}
\label{intro}
Uncertainty estimates are important across a wide range of applications, from medical diagnosis to weather forecasting to autonomous driving \cite{leibig2017leveraging,scher2018predicting,carvalho2015automated}. Accurately assessing the confidence of a prediction, and specifying the underlying distribution of potential errors, is a cornerstone of reliable and interpretable models. 
For example, having good prediction intervals when forecasting solar power production allows utilities to better account for fluctuations \cite{murata2018modeling}. Similarly, having reliable uncertainty estimates on a model assessing tumor size is important, as those metrics may be used to assess a variety of other clinical aspects \cite{kourou2015machine}. 

While uncertainty calibration for classification is a fairly well-developed research area, uncertainty calibration for regression has remained less explored and the techniques do not readily transfer \citep{kuleshov2018accurate}. Notably, previous work has indicated that the models which perform best on the regression tasks they are trained on will rarely be calibrated, and early stopping to guarantee calibration on a calibration dataset will usually hinder model performance overall \citep{laves2020wellcalibrated}.

\begin{figure}[t]
\begin{center}
\hspace{-0.1in}
\centerline{
\includegraphics[width=1.1\columnwidth]{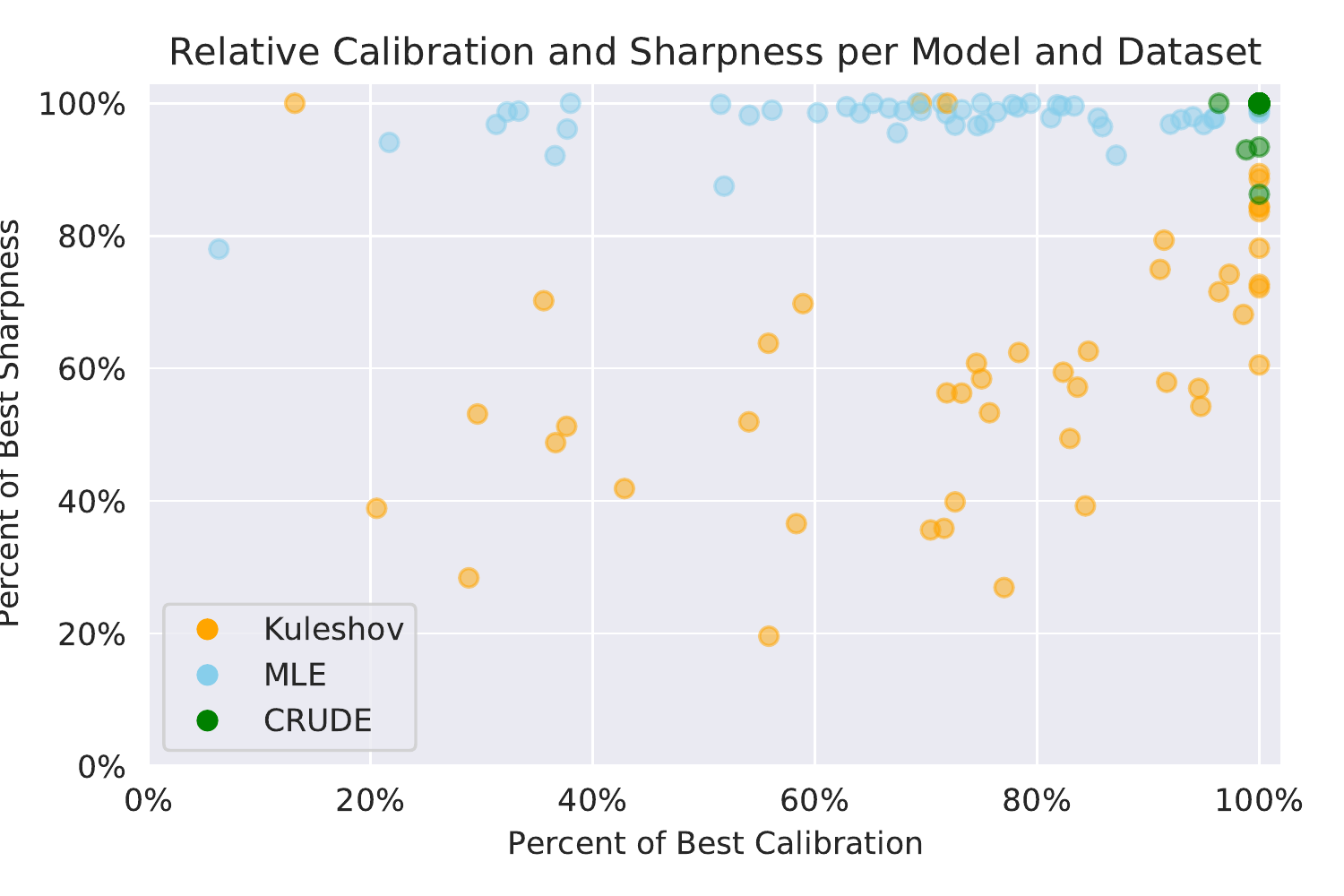}}
\caption{Comparison of calibration techniques, including CRUDE, a Gaussian MLE based on \citet{levi2019evaluating}, and the \citet{kuleshov2018accurate} method. Each point corresponds to the scores of a calibration technique on a particular machine learning model and UCI dataset. This figure visualizes each model's score relative to the best score on the dataset, so higher values correspond to better performance. 
}
\label{pull}
\end{center}
\end{figure}

Broadly, there is a trade-off discussed in the literature: some papers, like \citet{kuleshov2018accurate}, exclusively emphasize the calibration performance of calibration methods to produce well-calibrated uncertainty estimates, with little emphasis on the sharpness of the resulting methods. On the other hand, papers such as \citet{levi2019evaluating} emphasize simpler calibration techniques, trading some amount of calibration for a sharper model.

\vfill
\pagebreak
By generalizing key theoretical insights of \citet{kuleshov2018accurate}, \citet{levi2019evaluating}, and conformal inference methods \citep{barber2021predictive}, CRUDE recalibrates by leveraging the empirical distribution of prediction errors measured against a hold-out calibration set. This empirical distribution of errors then defines the \emph{shape} a family of distributions parameterized by a shift and scale value, which are obtained from the underlying model for each new prediction. 
We demonstrate, across a wide variety of datasets and models, that it is possible to have both state-of-the-art calibration and sharpness using CRUDE. 

\textbf{Contribution:} We propose a calibration method, \textit{Calibrating Regression Uncertainty Distributions Empirically} (CRUDE), inspired by prior work \cite{levi2019evaluating, kuleshov2018accurate, barber2021predictive}. CRUDE assumes less about the underlying error distribution and does not require an auxiliary calibration model, improving calibration and sharpness over both the \citeauthor{levi2019evaluating} and \citeauthor{kuleshov2018accurate} approaches on many datasets. Moreover, we demonstrate that this approach substantially improves the calibration of state-of-the-art probabilistic models for object detection. Furthermore, we demonstrate a direct mathematical equivalence with conformal inference, linking these two previously distinct approaches to accurate uncertainty estimates.

\begin{figure}[t]
\begin{center}
\hspace{-0.1in}
\centerline{
\includegraphics[width=1.1\columnwidth]{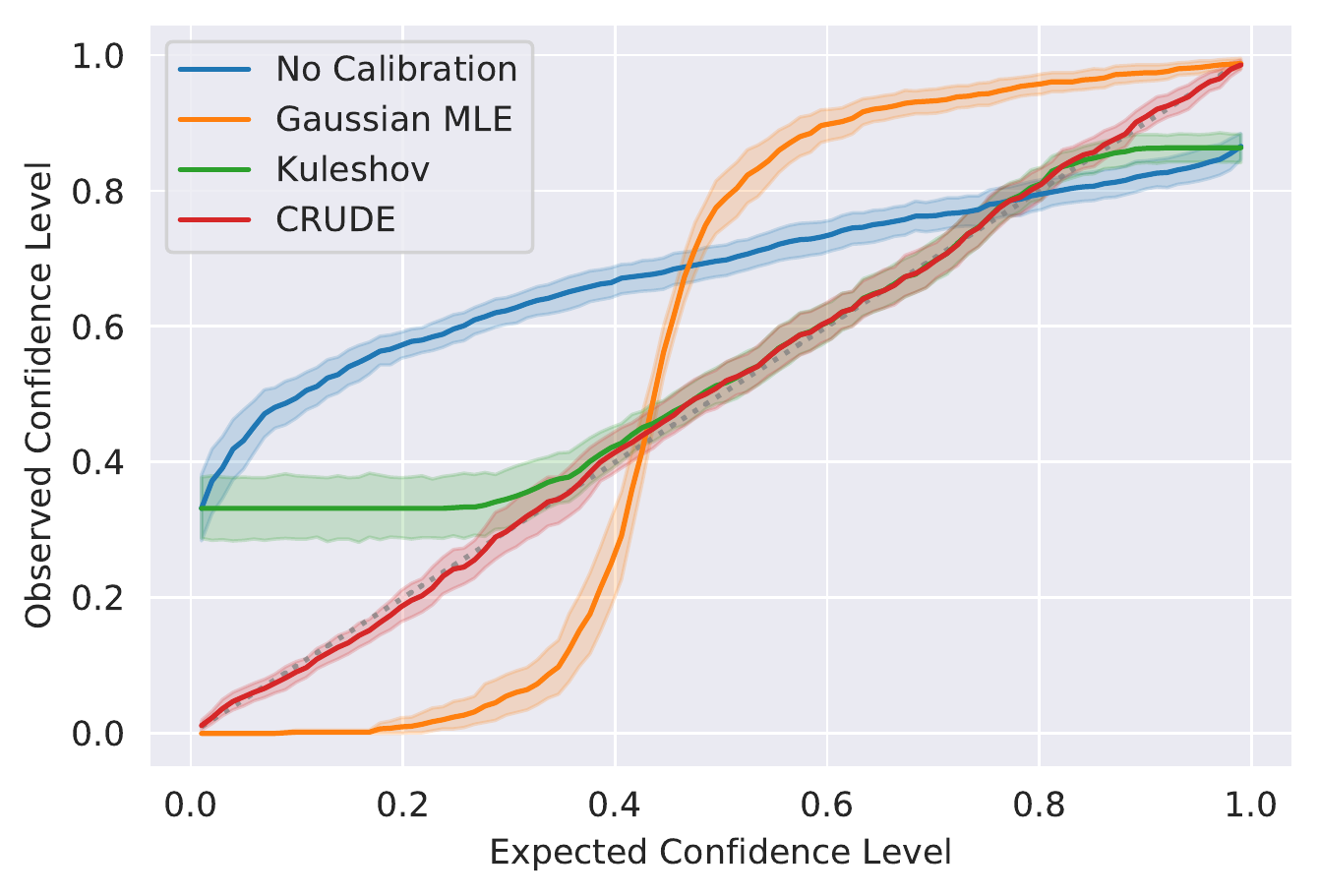}}
\caption{The calibration curves for calibrations of a neural network with a Monte Carlo dropout-derived uncertainty estimate. We show CRUDE, \citeauthor{kuleshov2018accurate}, and a Gaussian MLE based on \citeauthor{levi2019evaluating}, on the Forest Fires dataset \cite{Cortez2007ADM}. Note the ideal calibration curve is the line $x=y$, implying perfect calibration.
}
\label{auto}
\end{center}
\end{figure}

\vfill
\pagebreak

\section{Background}
\subsection{Regression Calibration}
Recently there has been an interest in uncertainty calibration in predictive models \citep{kumar2019verified, nixon2019measuring, schneider2020rethinking, maddox2019simple, zhu2019physics}. \citet{kuleshov2018accurate} proposed training an auxiliary model to calibrate uncertainty metrics to directly transform predicted uncertainties based on their associated probabilities on a calibration dataset. We note however that the definition in \citeauthor{kuleshov2018accurate} requires an invertible calibration curve, something that is typically missing from overconfident models, and often degrades the performance of the method.

\citet{levi2019evaluating} aimed to highlight theoretical issues with \citet{kuleshov2018accurate}, namely that the proposed algorithm tends to overfit and that their calibration metric allows for a model to be regarded as calibrated even when the calibrated uncertainties are uncorrelated with the true uncertainties. However, this criticism neglects the distinction between calibration and sharpness. Calibration evaluates the probabilistic accuracy of an uncertainty distribution, while sharpness rewards confident, correlated uncertainty predictions \citep{gneiting2007probabilistic}. \citeauthor{levi2019evaluating} proposed another calibration approach using maximum likelihood calibration over a normal distribution, measuring calibration using the absolute difference between the predicted uncertainties and the observed errors.

\citeauthor{levi2019evaluating} also notes that because the method recalibrates on the aggregate uncertainties, there exist probability distributions unrelated to the underlying distribution that can be found that correspond to ``perfect'' calibration regardless of the true uncertainty distributions.
Consequently, we show that the \citeauthor{kuleshov2018accurate} calibration method leads to less-sharp estimates of uncertainty. 
In theory, there are many transformations that lead to a calibrated distribution, but ideally we would like the one that results in the sharpest possible uncertainty estimates.

\begin{figure*}[t]
\begin{center}
\centerline{
\includegraphics[width=0.85\textwidth]{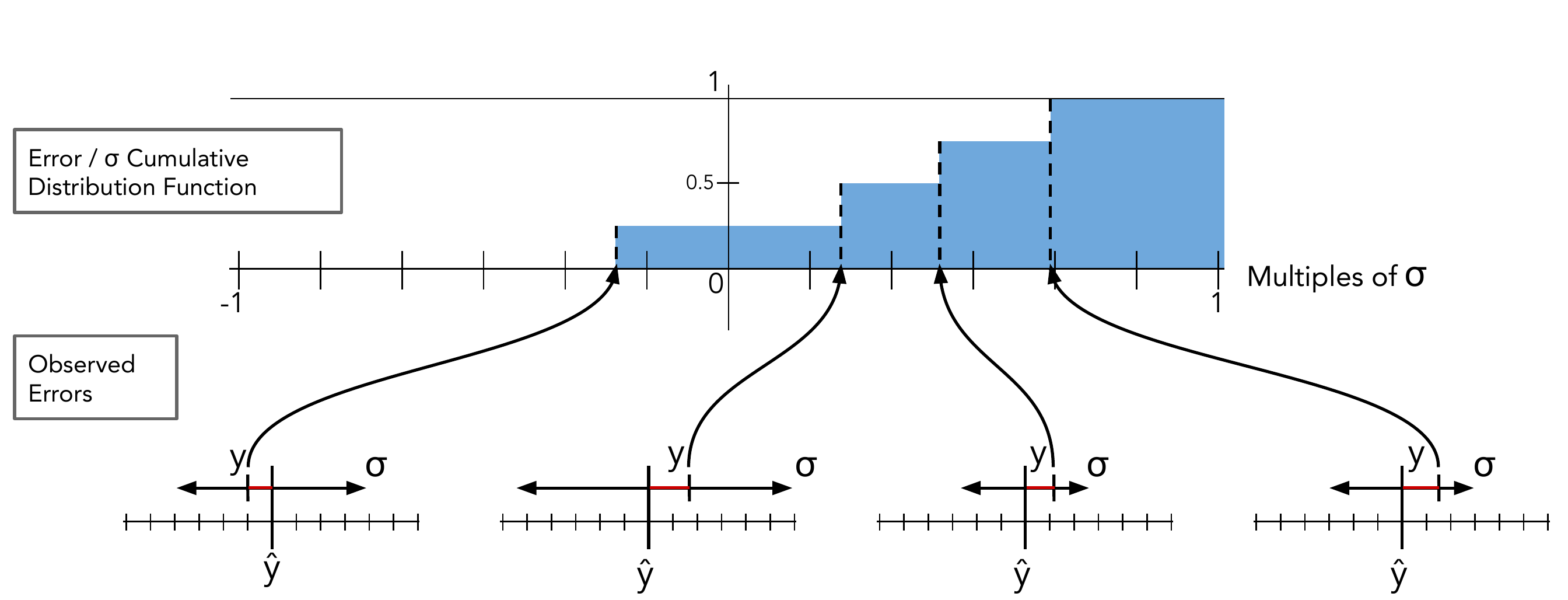}}
\caption{Visualization of the assumption made about the relationship between the underlying uncertainty distribution function and the observed errors: errors, scaled by $\sigma$, can be seen as samples from this underlying distribution function.}
\label{cdfvis}
\end{center}
\end{figure*}

\subsection{Aleatoric and Epistemic Uncertainty}
A recurring topic when discussing sources of uncertainty is the distinction between the two primary forms of uncertainty: aleatoric uncertainty corresponds to inherent uncertainty in the data (which \emph{cannot} be eliminated by collecting more data), while epistemic uncertainty, or knowledge uncertainty, corresponds to uncertainty derived from not having sufficient data in a certain region, typically manifested in the form of variance in the model parameters \citep{kendall2017uncertainties}. In addition, \citet{kendall2017uncertainties} discussed broadly the value of uncertainty estimates within computer vision, which motivates our analysis of the probabilistic object detection models in \citep{harakeh2021estimating}. 

As discussed in \citet{laves2020wellcalibrated}, aleatoric uncertainty can be captured by models which are trained to predict uncertainties in order to maximize likelihood. On the other hand, one context in which epistemic uncertainty can be measured is in analyzing the variance of a neural network's predictions with dropout \citep{laves2020wellcalibrated}. Within our work, we include models which are trained to anticipate both kinds of uncertainty in order to demonstrate its generality.

\subsection{Conformal Inference}

Since 2005, there has been a growing literature on conformal predictors, a broad class of interpretable probabilistic models for both regression and classification \citep{vovk2005conformal}. A conformal predictor uses as its central mechanism a nonconformity measure (in certain variations this measure is reduced to a function) which measures the similarity of one data point $(x_i, y_i)$ to a large pool of data, with the central motivation that an accurate prediction will tend to minimize this nonconformity. These scores are leveraged to create  confidence bounds (and implied posterior distributions) proportional to the empirically ranked conformity of data points. In the context of regression specifically, \citet{linusson2014signed} extended the utility of conformal inference with the introduction of negative nonconformity scores, which allowed for asymmetric posterior distributions (without any artificially asymmetric nonconformity scores). 

Recently, several methods have been developed which apply specific variants of conformal inference for regression. These variants have found applications in drug discovery \citep{cortes2020concepts} and biostatistics \citep{sun2017applying}, where their computational efficiency has been highlighted as a benefit, as well as in image classification where they are applied to neural networks \cite{matiz2019inductive}. However, conformal predictors in prior work do not incorporate the variance output from probabilistic models into their nonconformity measures: instead, as discussed in \citet{shafer2008tutorial}, methods that incorporate probabilistic models sample from predicted probabilistic distributions in order to get a more comprehensive set of predictions from which to calculate the nonconformity measure. 

In inductive or split conformal prediction \citep{papadopoulos2008inductive}, a dataset separate from the training or validation dataset is used as the basis from which to evaluate the nonconformity of a point on which a prediction is being made. Jackknife methods in conformal inference build on this idea, incorporating variation due to leave-one-out sampling for the conformity measure - in particular, the `jackknife+` method \citep{barber2021predictive} additionally incorporates variation due to the jackknife method in training. In CRUDE, these ideas are drawn upon and generalized.


\section{CRUDE}
\subsection{Derivation}
\label{deriv}

The CRUDE algorithm is designed for probabilistic regression problems, specifically for discriminative models that take input $x$ from space $\mathcal{X}$, and the output is a real valued $y \in \mathcal{R}$. The model, which we denote as $\mathcal{M}$, takes as input $x$, and outputs a pair of scale and non-negative shift parameters for each input $x$, i.e. $\mathcal{M}: \mathcal{X} \to \mathbb{R}\times\mathbb{R}_+$. We make no further assumptions about $\mathcal{M}$. Further, we assume that  each observed output is noisy, and the noise $z \in \mathbb{R}$ of each example is drawn independently and identically from an unspecified error distribution $\mathcal{E}$ over the real line. The observation $y$ itself is then just a scaled and shifted version of this noise, where $\sigma$ (scaling factor) and $\mu$ (shifting magnitude) are the parameters predicted by $\mathcal{M}$.

Based on these assumptions, we posit the following data generating process:

\begin{align*}
    x &\in \mathcal{X} \\
    \mu = \mu(x) &= \mathcal{M}_\text{shift}(x) \in \mathbb{R} \\
    \sigma = \sigma(x) &= \mathcal{M}_\text{scale}(x) \in \mathbb{R}_+ \\
    z &\sim \mathcal{E} \\
    y &= \mu + z\cdot\sigma
\end{align*}

It may be useful to think of $\mathcal{E}$ as a distribution over z-scores, whose samples are scaled and shifted (depending on $x$) to result in observations $y$. Further, note that in practice the model $\mathcal{M}$ is likely trained to predict the mean and standard deviation of a normal distribution (or some other distribution over the real line), though we simply interpret those predictions as  arbitrary scale and shift values per the data generation process described above, and discard any normality (or other distributional) assumptions.

The key objective of the CRUDE algorithm is to estimate $\mathcal{E}$, with $\hat{\mathcal{E}}$. This is done in a non-parametric way, simply by collecting sample z-scores from a hold-out calibration set $(X_C, Y_C)$, and constructing their empirical distribution:

\begin{equation*}
    Z_C = \left\{ \frac{y - \mu(x)}{\sigma(x)} \bigm\vert (x, y) \in (X_C, Y_C) \right\},
\end{equation*}
\begin{equation*}
    \hat{\mathcal{E}} := \text{Empirical}\left( Z_C\right).
\end{equation*}

The corresponding Cumulative Distribution Function (CDF), inverse CDF (Quantile function), and moments of $\hat{\mathcal{E}}$ are:

\begin{equation*}
    F_{\hat{\mathcal{E}}}(z) = \frac{\big\lvert\{z_c \le z \mid z_c \in Z_C\}\big\rvert}{|Z_C|},
\end{equation*}
\begin{equation*}
    F_{\hat{\mathcal{E}}}^{-1}(p) = \inf\left\{ z_c \in Z_C \mid F(z_c) \ge p \right\},
    \end{equation*}
    \begin{equation}
    \E\left[\hat{\mathcal{E}}\right] = \frac{1}{|Z_C|}\sum_{z_c \in Z_C} z_c, \label{eq:empmean}
    \end{equation}
    \begin{equation}
    \V\left[\hat{\mathcal{E}}\right] = \frac{1}{|Z_C|} \sum_{z_c \in Z_C} \left(z_c - \E\left[\hat{\mathcal{E}}\right]\right)^2. \label{eq:empvar}
\end{equation}

In turn, the predictive distribution on an unseen example $x_* \in \mathcal{X}$ is just the estimated $\hat{\mathcal{E}}$ scaled and shifted according to the predicted values $\mathcal{M}(x)$:

\begin{align*}
    y \mid x_* &\sim \mu(x_*) + \sigma(x_*)\cdot \hat{\mathcal{E}} \\
    &= \mathcal{M}_\text{shift}(x_*) + \mathcal{M}_\text{scale}(x_*)\cdot \hat{\mathcal{E}}.
\end{align*}

More concretely, the predicted distribution on an unseen example $x_*$ has the following distribution functions and moments:

\begin{equation*}
    F_{x_*}(y) = F_{\hat{\mathcal{E}}}\left(\frac{y - \mu(x_*)}{\sigma(x_*)}\right),
\end{equation*}
\begin{equation}
    F_{x_*}^{-1}(p) = \mu(x_*) + \sigma(x_*)\cdot F_{\hat{\mathcal{E}}}^{-1}(p), \label{eq:predquantile}
\end{equation}
\begin{equation}
    \E[y\mid x_*] = \mu(x_*) + \sigma(x_*)\cdot \E\left[\hat{\mathcal{E}}\right], \label{eq:predmean}
\end{equation}
\begin{equation}
    \V[y\mid x_*] = \sigma^2(x_*)\cdot \V\left[\hat{\mathcal{E}}\right]. \label{eq:predvar}
\end{equation}

Figure~\ref{cdfvis} provides a visual intuition of the empirical z-score distribution $\hat{\mathcal{E}}$.

\begin{table*}[t]
\caption{
\textbf{Performance on UCI Datasets.} Different calibration methods evaluated across various datasets. The sharpest overall calibrated models for each task are underlined, and sharpness of the uncalibrated distribution is reported in italics. Lower scores are better for both calibration and sharpness. The abbreviation Kule. refers to the \citet{kuleshov2018accurate} method. Note that the calibration error here is measured as the RMSE of the calibration curve rather than the MSE, for easier comparison.}
\label{results}
\begin{center}
\begin{small}
\begin{sc}
\resizebox{1.0\textwidth}{!}{%
\begin{tabular}{@{}l|rrrr|rrrr|rrrr|rrrl@{}}
\toprule
\multicolumn{1}{c}{} & \multicolumn{4}{c}{Variational NN Sharpness}                                                    & \multicolumn{4}{c}{Dropout NN Sharpness}                                                        & \multicolumn{4}{c}{NGBoost Sharpness}                                                           & \multicolumn{4}{c}{Gaussian Process Sharpness}                                                                      \\ \midrule
\multicolumn{1}{c}{} & \multicolumn{1}{c}{None} & \multicolumn{1}{c}{MLE} & Kule.       & \multicolumn{1}{c}{CRUDE} & \multicolumn{1}{c}{None} & \multicolumn{1}{c}{MLE} & Kule.       & \multicolumn{1}{c}{CRUDE} & \multicolumn{1}{c}{None} & \multicolumn{1}{c}{MLE} & Kule.       & \multicolumn{1}{c}{CRUDE} & \multicolumn{1}{c}{None} & \multicolumn{1}{c}{MLE} & Kule.             & \multicolumn{1}{c}{CRUDE} \\ \midrule
fire                 & \textit{0.190}           & 3.574                   & 9.646          & \textbf{3.435}            & \textit{0.124}           & 1.904                   & 9.597          & \textbf{1.879}            & \textit{0.165}           & 2.287                   & 8.224          & \textbf{2.214}            & \textit{0.940}           & 1.998                   & {\ul \textbf{1.840}} & 1.970                     \\
yacht                & \textit{0.101}           & 0.136                   & \textbf{0.119} & 0.128                     & \textit{0.154}           & 0.153                   & 0.167          & \textbf{0.141}            & \textit{0.011}           & 0.063                   & 0.158          & {\ul \textbf{0.062}}      & \textit{0.091}           & 0.127                   & 0.182                & \textbf{0.124}            \\
auto                 & \textit{0.174}           & 0.528                   & 0.819          & \textbf{0.511}            & \textit{0.122}           & 0.396                   & 0.773          & {\ul \textbf{0.382}}      & \textit{0.103}           & 0.536                   & 0.986          & \textbf{0.512}            & \textit{0.140}           & 0.411                   & 0.707                & \textbf{0.398}            \\
diabetes             & \textit{0.200}           & 1.472                   & 2.054          & \textbf{1.442}            & \textit{0.145}           & 0.856                   & 1.645          & {\ul \textbf{0.843}}      & \textit{0.263}           & 1.053                   & 1.472          & \textbf{1.027}            & \textit{0.234}           & 0.890                   & 1.364                & \textbf{0.870}            \\
housing              & \textit{0.138}           & 0.483                   & 1.172          & \textbf{0.467}            & \textit{0.146}           & 0.412                   & 1.110          & {\ul \textbf{0.398}}      & \textit{0.090}           & 0.501                   & 1.329          & \textbf{0.486}            & \textit{0.256}           & 0.535                   & 0.912                & \textbf{0.528}            \\
energy               & \textit{0.124}           & 0.140                   & 0.165          & \textbf{0.138}            & \textit{0.122}           & 0.263                   & 0.347          & \textbf{0.260}            & \textit{0.036}           & {\ul \textbf{0.059}}    & 0.070          & {\ul \textbf{0.059}}      & \textit{0.062}           & 0.084                   & 0.116                & \textbf{0.083}            \\
concrete             & \textit{0.170}           & 0.410                   & 0.712          & \textbf{0.407}            & \textit{0.196}           & 0.446                   & 0.818          & \textbf{0.436}            & \textit{0.187}           & 0.363                   & 0.623          & {\ul \textbf{0.355}}      & \textit{0.220}           & 0.510                   & 0.831                & \textbf{0.505}            \\
wine                 & \textit{0.462}           & 1.604                   & \textbf{1.585} & 1.593                     & \textit{0.129}           & 0.939                   & 2.405          & {\ul \textbf{0.935}}      & \textit{0.506}           & 0.951                   & 1.186          & \textbf{0.941}            & \textit{0.251}           & 1.079                   & 2.009                & \textbf{1.067}            \\
kin8nm               & \textit{0.305}           & 0.449                   & 0.754          & \textbf{0.448}            & \textit{0.182}           & 0.402                   & 0.958          & \textbf{0.401}            & \textit{0.561}           & 0.654                   & 0.737          & \textbf{0.653}            & \textit{0.191}           & {\ul \textbf{0.333}}    & 0.570                & {\ul \textbf{0.333}}      \\
power                & \textit{0.236}           & 0.243                   & 0.333          & \textbf{0.242}            & \textit{0.112}           & \textbf{0.283}          & 0.997          & \textbf{0.283}            & \textit{0.194}           & {\ul \textbf{0.236}}         & 0.390          & {\ul \textbf{0.236}}      & \textit{0.087}           & 1.490                   & \textbf{1.162}       & 1.347                     \\
airfoil              & \textit{0.352}           & 0.512                   & 0.686          & \textbf{0.509}            & \textit{0.170}           & 0.718                   & 1.445          & \textbf{0.705}            & \textit{0.304}           & 0.414                   & 0.526          & {\ul \textbf{0.411}}      & \textit{0.212}           & 0.547                   & 0.958                & \textbf{0.539}            \\
parkinsons           & \textit{0.121}           & \textbf{0.114}          & 0.158          & \textbf{0.114}            & \textit{0.114}           & 0.195                   & 0.310          & \textbf{0.194}            & \textit{0.094}           & {\ul \textbf{0.101}}    & 0.113          & {\ul \textbf{0.101}}      & \textit{0.142}           & 0.237                   & 0.411                & \textbf{0.223}            \\ \midrule
                     & \multicolumn{4}{c}{Variational NN Calibration}                                                  & \multicolumn{4}{c}{Dropout NN Calibration}                                                      & \multicolumn{4}{c}{NGBoost Calibration}                                                         & \multicolumn{4}{c}{Gaussian Process Calibration}                                                      \\ \midrule
                     & None                     & MLE                     & Kule.       & CRUDE                     & None                     & MLE                     & Kule.       & CRUDE                     & None                     & MLE                     & Kule.       & CRUDE                     & None                     & MLE                     & Kule.             & CRUDE                     \\ \midrule
fire                 & 0.154                    & 0.183                   & 0.098          & \textbf{0.069}            & 0.231                    & 0.192                   & 0.111          & \textbf{0.062}            & 0.188                    & 0.182                   & 0.074          & \textbf{0.057}            & 0.221                    & 0.224                   & 0.114                & \textbf{0.082}            \\
yacht                & 0.144                    & 0.164                   & \textbf{0.085} & 0.086                     & 0.121                    & 0.101                   & \textbf{0.088} & \textbf{0.088}            & 0.167                    & 0.135                   & 0.115          & \textbf{0.097}            & 0.097                    & 0.072                   & 0.070                & \textbf{0.069}            \\
auto                 & 0.155                    & 0.080                   & 0.097          & \textbf{0.076}            & 0.150                    & 0.085                   & 0.088          & \textbf{0.073}            & 0.186                    & 0.089                   & 0.111          & \textbf{0.060}            & 0.162                    & 0.075                   & 0.096                & \textbf{0.069}            \\
diabetes             & 0.228                    & 0.067                   & 0.177          & \textbf{0.063}            & 0.221                    & \textbf{0.061}          & 0.162          & \textbf{0.061}            & 0.187                    & 0.071                   & 0.112          & \textbf{0.066}            & 0.198                    & 0.075                   & 0.129                & \textbf{0.072}            \\
housing              & 0.148                    & 0.084                   & 0.084          & \textbf{0.061}            & 0.139                    & 0.071                   & 0.074          & \textbf{0.053}            & 0.201                    & 0.093                   & 0.120          & \textbf{0.070}            & 0.079                    & 0.072                   & 0.060                & \textbf{0.055}            \\
energy               & 0.069                    & 0.083                   & \textbf{0.050} & \textbf{0.050}            & 0.086                    & 0.075                   & 0.056          & \textbf{0.051}            & 0.070                    & 0.068                   & \textbf{0.054} & \textbf{0.054}            & 0.064                    & \textbf{0.053}          & 0.055                & 0.055                     \\
concrete             & 0.094                    & 0.069                   & 0.055          & \textbf{0.046}            & 0.133                    & 0.062                   & 0.070          & \textbf{0.053}            & 0.092                    & 0.064                   & 0.055          & \textbf{0.052}            & 0.110                    & 0.056                   & 0.055                & \textbf{0.041}            \\
wine                 & 0.101                    & 0.096                   & 0.046          & \textbf{0.032}            & 0.240                    & 0.045                   & 0.180          & \textbf{0.037}            & 0.078                    & 0.057                   & 0.035          & \textbf{0.032}            & 0.182                    & 0.046                   & 0.108                & \textbf{0.032}            \\
kin8nm               & 0.066                    & 0.018                   & 0.017          & \textbf{0.014}            & 0.121                    & 0.022                   & 0.042          & \textbf{0.018}            & 0.052                    & 0.033                   & \textbf{0.017} & \textbf{0.017}            & 0.079                    & 0.024                   & 0.024                & \textbf{0.018}            \\
power                & 0.016                    & 0.018                   & \textbf{0.015} & \textbf{0.015}            & 0.138                    & 0.021                   & 0.052          & \textbf{0.015}            & 0.024                    & 0.023                   & \textbf{0.015} & \textbf{0.015}            & 0.160                    & 0.189                   & 0.091                & \textbf{0.012}            \\
airfoil              & 0.059                    & 0.046                   & 0.037          & \textbf{0.036}            & 0.158                    & 0.061                   & 0.090          & \textbf{0.033}            & 0.059                    & \textbf{0.039}          & \textbf{0.039} & \textbf{0.039}            & 0.105                    & 0.064                   & 0.056                & \textbf{0.041}            \\
parkinsons           & 0.060                    & 0.050                   & \textbf{0.019} & \textbf{0.019}            & 0.071                    & 0.035                   & 0.026          & \textbf{0.022}            & 0.035                    & 0.026                   & \textbf{0.018} & \textbf{0.018}            & 0.028                    & 0.083                   & 0.019                & \textbf{0.018}            \\ \bottomrule
\end{tabular}%
}

\end{sc}
\end{small}
\end{center}
\end{table*}

\subsection{Algorithm}

The CRUDE algorithm assumes that there exists a model $\mathcal{M} = (\mathcal{M}_\text{shift}, \mathcal{M}_\text{scale}) = (\mu, \sigma)$ already trained on the training data. The algorithm treats the model $\mathcal{M}$ as a given black-box, and does not require further access to the training data. Instead, an independent hold-out calibration set $(X_C, Y_C) \in (\mathcal{X} \times \mathbb{R})^L$ of size $L \in \mathbb{N}$  is used calibrate the model. On this hold-out calibration set we first calculate the residuals between the observed labels $y$ and the predicted shift value $\mu(x)$, and inversely scale by the predicted scale $\sigma(x)$. These z-scores are sorted and the empirical distribution $\hat{\mathcal{E}}$ formed by them is considered to be our estimate of $\mathcal{E}$. With this constructed $\hat{\mathcal{E}}$ in hand, we are now able to answer probabilistic questions about the predictive distribution for any unseen example input $x_*$.



\begin{algorithm}[h]
  \caption{CRUDE calibration algorithm, predicting the calibrated $p^{th}$ quantile for input $x_*$}
  \label{alg:crude}
  \KwData{Held-out calibration dataset $(X_C, Y_C) \in (\mathcal{X}\times \mathbb{R})^{L}$ }
  \KwIn{ Test example $x_{*}$, Target quantile level $p$, and Model $\{\mu, \sigma\} : \mathcal{X} \rightarrow (\mathbb{R},\mathbb{R}_+)$}
  \KwOut{ Target quantile $q$}
  $Z_C \gets [\quad ]$

  \For{$(x,y) \gets (X_C, Y_C)$}{
    $Z_{C} \leftarrow \text{SortedInsert}\left\{Z_C, \frac{y - \mu(x)}{\sigma(x)}\right\}$
  } 
  $z_p \gets Z_C\left[\text{int}(p \cdot L)\right]$

  $q \gets \mu(x_*) + \sigma(x_*) \cdot z_p$
\end{algorithm}

 In Algorithm~\ref{alg:crude} we outline the steps to make a quantile prediction at a given level $p$, using which we can construct prediction intervals. For example, to calculate a $90^{th}$ percentile prediction interval of $y$ for a given input $x_*$, we invoke Algorithm~\ref{alg:crude} twice: once with $p=0.05$ and again with $p=0.95$ to obtain the lower and upper bounds of the interval respectively. Note that the construction of the sorted $Z$ can be cached in the first invocation and re-used in future queries. This results in a preprocessing run-time complexity of $O(L\log L)$ for the first time, and $O(1)$ time for all future quantile queries. 
 
 Similarly, querying the moments of the predictive distribution involves a one-time preprocessing time complexity of order $O(L)$ (to calculate Equation~\ref{eq:empmean} or Equation~\ref{eq:empvar}), followed by a $O(1)$ time complexity for all future queries (Equation~\ref{eq:predmean} or Equation~\ref{eq:predvar}).
 
 In this time complexity analysis we only focus on the steps required by CRUDE, and ignore the time taken by the black-box model in computing $\mu(x_*)$ and $\sigma(x_*)$. 

\subsection{Relationship to the Conformal Framework}

In addition to the posited data generative process in Sec 3.1, the CRUDE algorithm can also be understood as a variant of the Inductive Conformal Prediction (ICP) framework \cite{linusson2014signed}.   

A conformal predictor utilizes nonconformity measure  
\begin{equation}
A: B \times (x,y) \rightarrow \mathbb{R}
\end{equation}
where $B$ represents a bag of exchangeable examples, and $(x, y)$ is a new example whose non-conformance is being measured w.r.t $B$. In our case, $B$ the hold-out calibration dataset of size $L$: 

\begin{equation*}
 B = (X_C, Y_C) \in (\mathcal{X} \times \mathbb{R})^L.
\end{equation*}

Conventional nonconformity measures have non-negative outputs, and such measures are typically utilized for generating two sided confidence bounds, for some confidence level $p$ in the output space. In the case of ICP, such confidence bounds correspond to prediction intervals over $y$ for a given input $x_*$. The calibration set $B$ is used to determine an appropriate $z_p$ corresponding to the desired confidence bound $p$ (e.g. $p=90\%$), such that $z_p$ best satisfies the condition

\begin{equation}
\cfrac{\big\lvert\{(x, y) \in B \mid A(B, (x, y)) < z_p \}\big\rvert}{|B|} \approx p.
\label{eq:seven}
\end{equation} 

Then, the prediction interval $y^{[p]}|x_*$ over $y$ for given input $x_*$, and desired confidence level $p$ is:
\begin{equation}
y^{[p]}|x_* = \{y\in \mathbb{R} \mid A (B, (x_*, y)) \leq z_p\}, 
\label{eq:six}
\end{equation}
where $z_p$ is according from Equation~\ref{eq:seven}. 

Traditionally nonconformity measures $A$ are non-negative, where the above formulation of generating predictive internals necessarily produces symmetric bounds. \citet{linusson2014signed}'s extension of this method allows for one sided bounding (and thus, a naturally asymmetric CDF) in a regression context by utilizing a nonconformity score that allows for negative values, with a value of $0$ as the maximum possible conformity. This in turn allows the specification of desired independent upper and lower bounds for prediction, $p_l$ and $p_u$ (eg. $p_l=5\%, p_u=95\%$ for an overall confidence bound of 90\%). This in turn results in corresponding $z_l$ and $z_u$ as calculated from Equation~\ref{eq:seven}, which are then used to generate the desired prediction interval, denoted $y^{[l,u]}|x_*$:

\begin{equation} \label{eq:twosided}
    y^{[l,u]}|x_* = \{y\in \mathbb{R} \mid z_l \leq A (B, (x_*, y)) \leq z_u\}.
\end{equation} 

\begin{figure*}[t]
\begin{center}
\hspace{-5px}
\centerline{\includegraphics[width=0.9\textwidth]{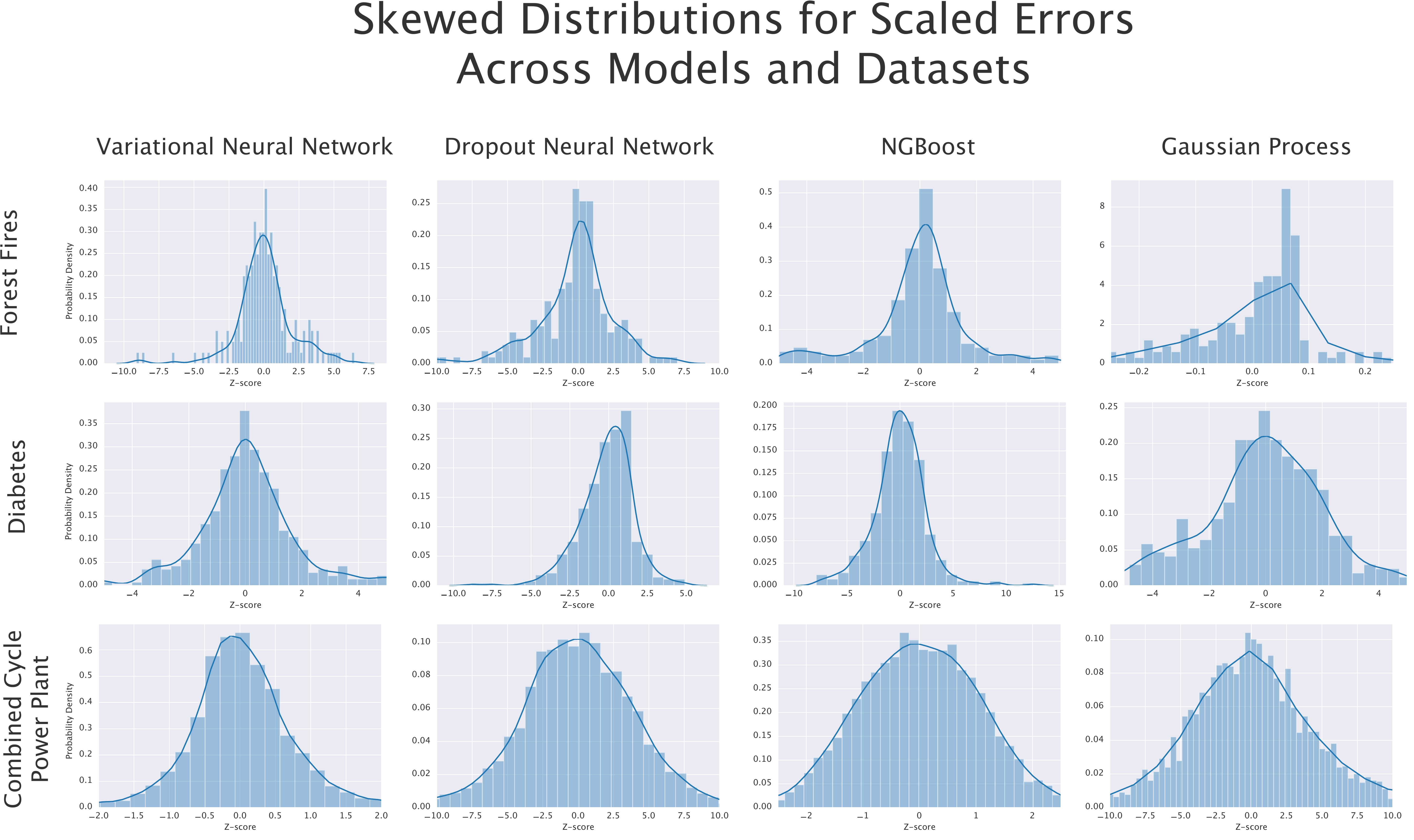}}
\caption{Visualization of the distribution of the scaled errors for each model on the Fores Fires dataset, the Diabetes dataset, and the Combined Cycle Power Plant dataset \citep{Cortez2007ADM,kahn1993diabetes,tufekci2014prediction}; each distribution has a skew or kurtosis. 
}
\vspace{-20px}
\label{skew}
\end{center}
\end{figure*}

To see the connection to CRUDE, consider a given probabilistic model $\mathcal{M} = (\mu, \sigma)$, and a nonconformity score defined as:

\begin{equation} 
\begin{aligned}
    A_\mathcal{M}\left(B, (x, y)\right) = \cfrac{y - \mu(x)}{\sigma(x)}
\end{aligned}
\end{equation}

Next, note that Equation~\ref{eq:seven} can be rewritten, given a sufficiently large calibration data $B$ of size $L$, as 
\begin{equation*}
\begin{aligned}
    z_p & \approx \text{percentile}\left(\left[A_\mathcal{M}(B, (x,y)) \mid  (x, y)  \in B \right], p\right) \\ 
    & = \text{percentile}\left(\left\{\cfrac{y - \mu(x)}{\sigma(x)} \mid (x, y) \in B \right\}, p\right) \\
    &= \text{Sorted}\left\{\cfrac{y - \mu(x)}{\sigma(x)} \mid (x, y) \in B \right\} \left[\text{int}( p \cdot L)\right]
\end{aligned}
\end{equation*}
Note that $z_p$ here corresponds directly to  $z_p$ in Algorithm~\ref{alg:crude}. Now consider the prediction interval generated by CRUDE for an equivalent $p_l$ and $p_u$ as above. From Algorithm~\ref{alg:crude} we can see that:

\begin{align*}
    y^{[l,u]}|x_* &= \{y \in \mathbb{R} \mid \text{Algo1}(p_l) \le y \le \text{Algo1}(p_u) \} \\
    &= \{ y \in \mathbb{R} \mid \mu(x_*) + \sigma(x_*)\cdot z_l \le y \\&\quad\quad\quad\quad\quad\quad \le  \mu(x_*) + \sigma(x_*)\cdot z_u \} \\
    &= \left\{ y \in \mathbb{R} \mid z_l \le \left(\cfrac{y - \mu(x_*)}{\sigma(x_*)} \right) \le z_u \right\} \\
    &= \left\{ y \in \mathbb{R} \mid z_l \le A_\mathcal{M}(B, (x_*, y) \le z_u \right\} \\
    &= \text{Equation}~\ref{eq:twosided}.
\end{align*}

\section{Evaluating Calibration and Sharpness}
\label{calsharp}

\textbf{Calibration.} Because we can associate each error with a likelihood in a regression prediction with uncertainty, we can evaluate the relationship between the expected and observed confidence levels. That is, for a calibrated model, the errors it predicts will be in the $30^{th}$ percentile of its distribution, should be above the observed errors 30\% of the time. This is the probability integral transform (PIT) value and well-calibrated models have a uniform distribution over the percentiles associated with errors \cite{gneiting2007probabilistic}. 

Ideally, each expected confidence percentile should match the fraction of values observed below the predictions of that percentile. To measure calibration, we use a variant of the \citeauthor{kuleshov2018accurate} metric, measuring the RMSE between the expected confidence levels and the observed confidence levels. We use the RMSE rather than the MSE in order to more directly reflect the percent error, but the model rankings from the two approaches are unchanged. Let the test set against which we are measuring calibration be denoted $D_T \in (\mathcal{X}\times\mathbb{R})^N$ of size $N$. We initialize values of $p$ in the range $[0,1]$ with step size\footnote{As the calibration score corresponds to the root-mean-squared error (RMSE) to the ideal calibration curve, and the calibration curve is monotonically increasing, the calibration score converges as the step size decreases.}
of $1/S$ where $S\in\mathbb{N}$ (with $S=100$ in our experiments), and for each $p_j$ find empirical frequency $\hat{p}_j$: 

\begin{equation}
    \hat{p}_j = \frac{1}{|D_T|} \sum_{(x,y) \in D_T} \mathbf{1}\left\{ y < F_x^{-1}(p_j) \right\}, \label{calibrationscore}
\end{equation}

where $F_x^{-1}$ is the predictive quantile function corresponding to the calibration method being evaluated. In case of CRUDE, the predictive quantile function is defined in Equation~\ref{eq:predquantile}.

The overall calibration score across all the $S$ comparison points is then:
\begin{equation}
    \text{cal}(\hat{p}, p) =   \sqrt{\frac{1}{S} \sum_{j=0}^{S}(\hat{p}_{j} - p_j)^2}.
    \label{calibrationscore2}
\end{equation}

\textbf{Sharpness.} Calibration does not tell us the full story: a calibration method's efficacy on a dataset is also based on the resulting sharpness. 
In order to evaluate sharpness, we can use the mean of the calibrated predicted variance on the validation set \cite{gneiting2007probabilistic}, and we take the square root of this value to match the error's dimensionality. Note that a lower score implies higher sharpness.

\section{Experiments}
\subsection{UCI Dataset Experiments}
We evaluate each calibration method on multiple probabilistic predictive models which have shown good performance on numerous tasks, including two flavors of Bayesian neural networks, Gaussian processes, and Natural Gradient Boosting \cite{duan2019ngboost}. Specifically, the two neural network approaches that we consider include one using Monte Carlo dropout as described in \citet{gal2015dropout} as well as one which separately predicts a mean and variance \citep{papadopoulos2001confidence}, both with hyperparameters similar to those described in \citet{gal2015dropout}. Further detail about the training of these models and their architectures is provided in Appendix~\ref{modelinfo}. 

We compare no calibration, a Gaussian maximum likelihood estimate (including shift) based on \citet{levi2019evaluating}, the \citet{kuleshov2018accurate} method, and CRUDE. We consider 12 datasets, mostly from the public Machine Learning Repository from the University of California, Irvine. We chose these datasets to include a variety of applications and technical features, including input dimensionality and dataset size, as well as the downstream utility of calibrated uncertainties for these predictions. We include more detail about the UCI datasets in Appendix~\ref{datasetinfo}.

\subsubsection*{Evaluation}
For each model on each dataset, we run 20 trials with the dataset shuffled and split repeatedly, with a $(0.5, 0.4, 0.1)$ split between training, calibration, and validation data. For each trial, we score every calibration method for calibration and sharpness. We use the metrics discussed in Section~\ref{calsharp} for calibration and sharpness. Note that we do not use the calibration metrics proposed in \citet{levi2019evaluating,laves2020wellcalibrated}, as they do not consider the uncertainty distribution associated with a given uncertainty estimate, instead using mean absolute error as a metric. Moreover, as highlighted in the appendix of \citeauthor{laves2020wellcalibrated}, the metric implicitly corresponds to log-likelihood of the observations under a Laplacian prior, which does not match the prior that most uncertainty models are trained under.

\begin{table}[t]
\caption{
\textbf{Bounding Box Results.} Different calibration methods evaluated across various pretrained models from \citeauthor{harakeh2021estimating}. Note again that the calibration error here is measured as the RMSE of the calibration curve rather than the MSE, for easier comparison. In addition, the calibration of the pre-trained models varies slightly from what was originally published in \citet{harakeh2021estimating}; this is due to an error in evaluation in the code base for the original paper, which has since been corrected.}
\label{results_2}
\begin{center}
\begin{small}
\begin{sc} 
\resizebox{0.48\textwidth}{!}{%
\begin{tabular}{@{}clllll@{}}
\multicolumn{1}{l}{}        &      & \multicolumn{4}{c}{Sharpness}                                    \\ \cmidrule(l){3-6} 
Detector                    & Loss & None  & Gauss. & Kule.       & CRUDE          \\ \midrule
\multirow{3}{*}{DETR}       & DMM  & \textit{0.004} & 0.113        & 0.265          & \textbf{0.110} \\
                            & ES   & \textit{0.014} & 0.132        & 0.203          & \textbf{0.130} \\
                            & NLL  & \textit{0.075} & 0.084        & \textbf{0.066} & 0.081          \\ \midrule
\multirow{3}{*}{RetinaNet}  & DMM  & \textit{0.011} & 0.125        & 0.241          & \textbf{0.109} \\
                            & ES   & \textit{0.016} & 0.109        & 0.198          & \textbf{0.097} \\
                            & NLL  & \textit{0.049} & 0.108        & \textbf{0.096} & \textbf{0.096} \\ \midrule
\multirow{3}{*}{FasterRCNN} & DMM  & \textit{0.008} & 0.090        & 0.217          & \textbf{0.083} \\
                            & ES   & \textit{0.013} & 0.088        & 0.171          & \textbf{0.081} \\
                            & NLL  & \textit{0.016} & 0.090        & 0.169          & \textbf{0.085} \\ \midrule
\multicolumn{1}{l}{}        &      & \multicolumn{4}{c}{Calibration}                                  \\ \cmidrule(l){3-6} 
\multirow{3}{*}{DETR}       & DMM  & {0.136}  & 0.172        & 0.074          & \textbf{0.011} \\
                            & ES   & {0.111}  & 0.161        & 0.040           & \textbf{0.011} \\
                            & NLL  & {0.164}  & 0.159        & \textbf{0.012} & \textbf{0.012} \\ \midrule
\multirow{3}{*}{RetinaNet}  & DMM  & {0.184}  & 0.113        & 0.091          & \textbf{0.012} \\
                            & ES   & {0.147}  & 0.112        & 0.058          & \textbf{0.012} \\
                            & NLL  & {0.082}  & 0.109        & \textbf{0.016} & \textbf{0.012} \\ \midrule
\multirow{3}{*}{FasterRCNN} & DMM  & {0.147}  & 0.129        & 0.065          & \textbf{0.012} \\
                            & ES   & {0.105}  & 0.127        & 0.043          & \textbf{0.011} \\
                            & NLL  & {0.101}  & 0.128        & 0.036          & \textbf{0.013} \\ \bottomrule
\end{tabular}%
}
\end{sc}
\end{small}
\end{center}
\end{table}

\vfill
\pagebreak
\subsection{Calibrated Object Detection}
Following recent work on the benefits of regression uncertainty in deep object detection, we demonstrate that CRUDE can substantially improve the calibration of predictions in state-of-the-art probabilistic object detection models \citep{harakeh2021estimating}. While probabilistic estimates carry some inherent benefits, discussed in \citet{kendall2017uncertainties}, \citet{harakeh2021estimating} presented a probabilistic extension of a deterministic computer vision model which outperformed its corresponding deterministic model on mean Average Precision (mAP), a non-probabilistic object detection metric. 

\begin{figure}[t]
\begin{center}
\hspace{-0.2in}
\centerline{
\includegraphics[width=0.5\textwidth]{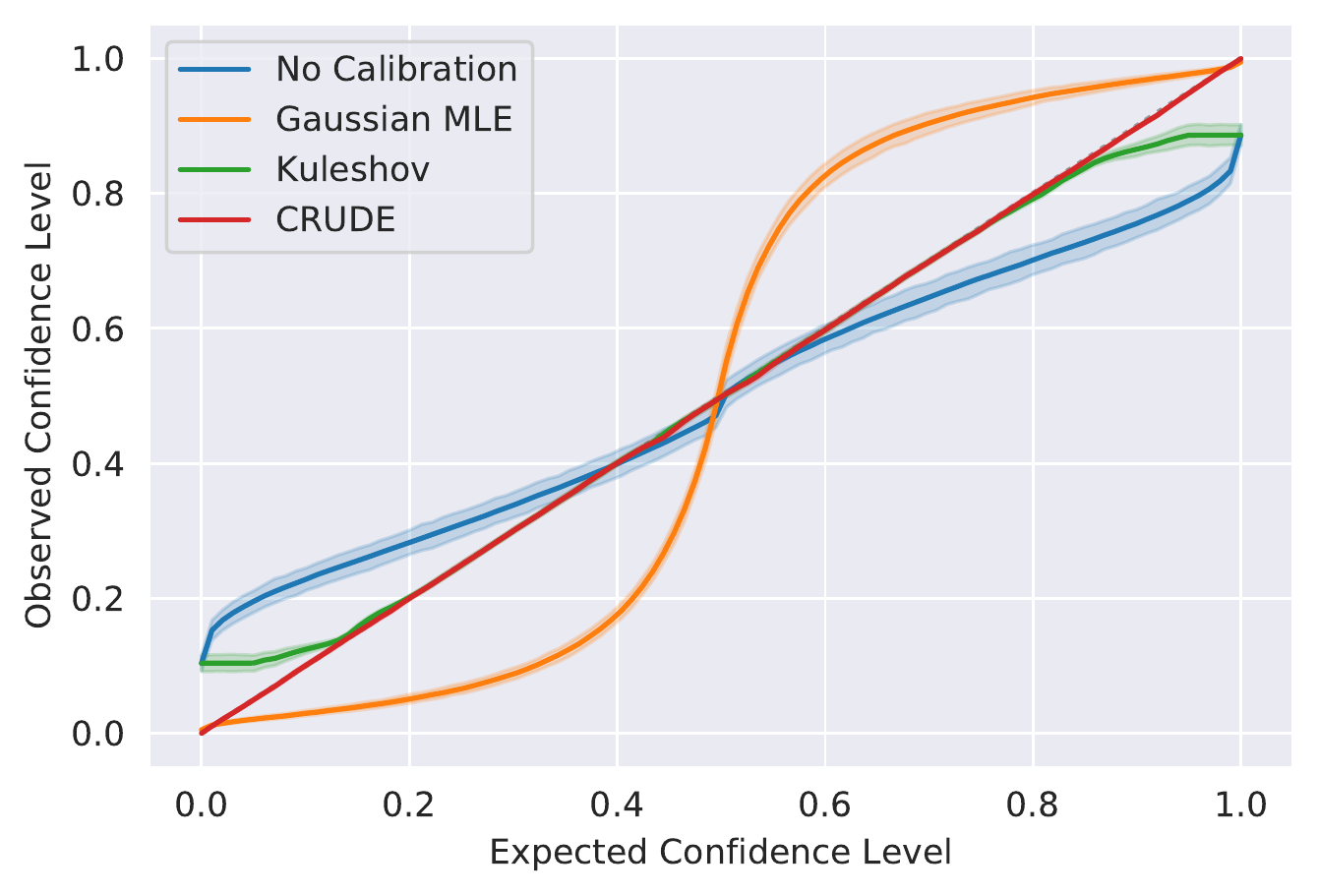}}
\caption{The calibration curves for the DETR model with an energy-based loss on the COCO \citep{lin2014microsoft} dataset, which was the best-performing model in \citet{harakeh2021estimating} in terms of mAP on both the COCO \citep{lin2014microsoft} and OpenImages \citep{kuznetsova2020open} datasets.}
\label{calcurve}
\end{center}
\end{figure}

However, as with many probabilistic models, the best-performing model from the \citet{harakeh2021estimating} work, in terms of mAP, was poorly calibrated, with a RMSE of 11.1\% between the ideal calibration curve and the measured calibration curve, shown in Figure~\ref{calcurve}. We note that one of the benefits of a post-hoc calibration method is the ability to apply it to pre-trained models. Specifically, we calibrate all nine of \citet{harakeh2021estimating}'s pretrained models for a bounding box regression task on the COCO \citep{lin2014microsoft} data, using each combination of three detectors and three loss functions. 

While \citet{harakeh2021estimating} fit a Gaussian posterior to the data, we simply use the predicted mean and variance out of context to re-estimate a complete posterior using CRUDE.  We split a non-repeating sample of 5,000 predicted bounding boxes from the validation set in half, and use one half for calibration and the other half to evaluate the calibration RMSE. Once again, we compare to a Gaussian MLE and the \citet{kuleshov2018accurate} method.

\section{Discussion}
As highlighted in Table~\ref{results}, we find that on the decisive majority of UCI dataset \citep{dua2019} tasks and models, our calibration method outperforms the alternative methods in both calibration and sharpness. 
There is no machine learning model that is consistently the sharpest across all datasets when calibrated, though the dropout neural network and NGBoost are disproportionately represented among the sharpest solutions. 


In general, we can delineate the primary theoretical sources of error for CRUDE: the assumption that the error distributions are identical may be wrong, the assumption that the error distributions are independent may be wrong, and there may not be enough data to accurately estimate $\hat{\mathcal{D}}$. In addition, the Gaussian maximum likelihood estimation technique may outperform CRUDE if the errors are truly perfectly Gaussian, though our results indicate that in practice this is rarely the case. 

We notice that the \citeauthor{kuleshov2018accurate} method has the greatest likelihood of matching CRUDE on the NGBoost model in terms of calibration on larger datasets, but this tendency is not replicated for the other models. However, in almost all cases where the \citeauthor{kuleshov2018accurate} method results in equal calibration, there is a substantial sacrifice of sharpness. As highlighted by \citeauthor{levi2019evaluating}, \citeauthor{kuleshov2018accurate} will be able to calibrate any distribution with enough data (excluding a very overconfident model), but it will often do so at the expense of sharpness. This is visualized by Figure~\ref{pull}, which shows that Kuleshov is well-calibrated, but rarely sharp, while the Gaussian MLE is often sharp, but usually more-poorly calibrated.
Figure~\ref{skew} shows an example scaled-error distribution on several datasets, indicating that the skew and kurtosis of the underlying error distributions may help to explain much of the improvement \citep{tufekci2014prediction}. 

On the computer vision task, as highlighted in Table~\ref{results_2} as well as in Figure~\ref{calcurve}, CRUDE is consistently the best-calibrated approach, and the sharpest calibrated approach on all but one model. On the best-performing model in terms of mAP, the DETR model with an energy-based loss, the calibration RMSE is reduced by 90\% relative to the uncalibrated model, and 72\% relative to Kuleshov \cite{kuleshov2018accurate}.

\vfill
\pagebreak

\section{Conclusion and Future Directions}
CRUDE offers substantial improvements over existing regression calibration techniques, especially for datasets where the error distribution has a nonstandard shape. However, there are many meaningful avenues left to explore. For most datasets, the assumption of a fixed uncertainty distribution, whether Gaussian or empirical, is incorrect to varying degrees. There are many ways to extend CRUDE to account for uncertainty distributions varying with respect to inputs, such as by calculating the percent-point function using only a given input's nearest neighbors. 

Similarly, treating output dimensions as independent is inadequate for many models, especially those producing a full covariance matrix as their uncertainty estimate. While this may be solvable with the inverse of pseudo-inverse of the predicted covariance matrix in place of dividing by the uncertainty in each dimension, the question of selecting a percentile over a higher-dimensional list of vectors is non-trivial. The successful incorporation of distinct epistemic and aleatoric uncertainty measurements in the calibrated distribution would be an exciting extension of CRUDE.   
Additionally, it may be possible to leverage the richer distributional information provided by Monte Carlo dropout. Ultimately, CRUDE opens the door to many improved regression calibration methods. 


\vfill
\pagebreak
{
\small{
\bibliography{paper}
}
}
\vfill

\appendix
\vspace{-10px}
\pagebreak
\onecolumn
\section*{Appendix}

\section{Model Training Information}
\label{modelinfo}
The Gaussian processes are optimized using GPytorch with an RBF kernel and learning rate of $0.01$ \cite{gardner2018gpytorch}. The neural networks we evaluate use three hidden layers with 1024 units each, learning rate $0.0001$, and weight decay$=0.01$. The neural networks are trained with the Adam optimizer \cite{kingma2014adam}, with $p=0.2$ if using dropout.

\section{UCI Dataset Information}
\label{datasetinfo}
We analyze the Forest Fires, Yacht Hydrodynamics, Auto MPG, Diabetes, Boston Housing, Energy Efficiency, Concrete Compressive, Wine Quality, Combined Cycle Power Plant, Airfoil Self-Noise, kin8nm (not UCI), and Parkinsons Telemonitoring datasets \citep{Cortez2007ADM,ortigosa2007neural,autompg,kahn1993diabetes,harrison1978hedonic,tsanas2012accurate,yeh1998modeling,cortez2009modeling,tufekci2014prediction,brooks1989airfoil,ghahramani1996kin,tsanas2009accurate}. The inputs range from 4 dimensional in the Power Efficiency dataset to 21 dimensional in the Parkinson's dataset, with a median 8 dimensions. The sample count ranges from 308 examples in the Yacht Hydrodynamics dataset to 9568 examples in the Combined Cycle Power Plant dataset, with a median 899 examples.

\begin{figure}[h]
\centerline{\includegraphics[width=0.39\textwidth]{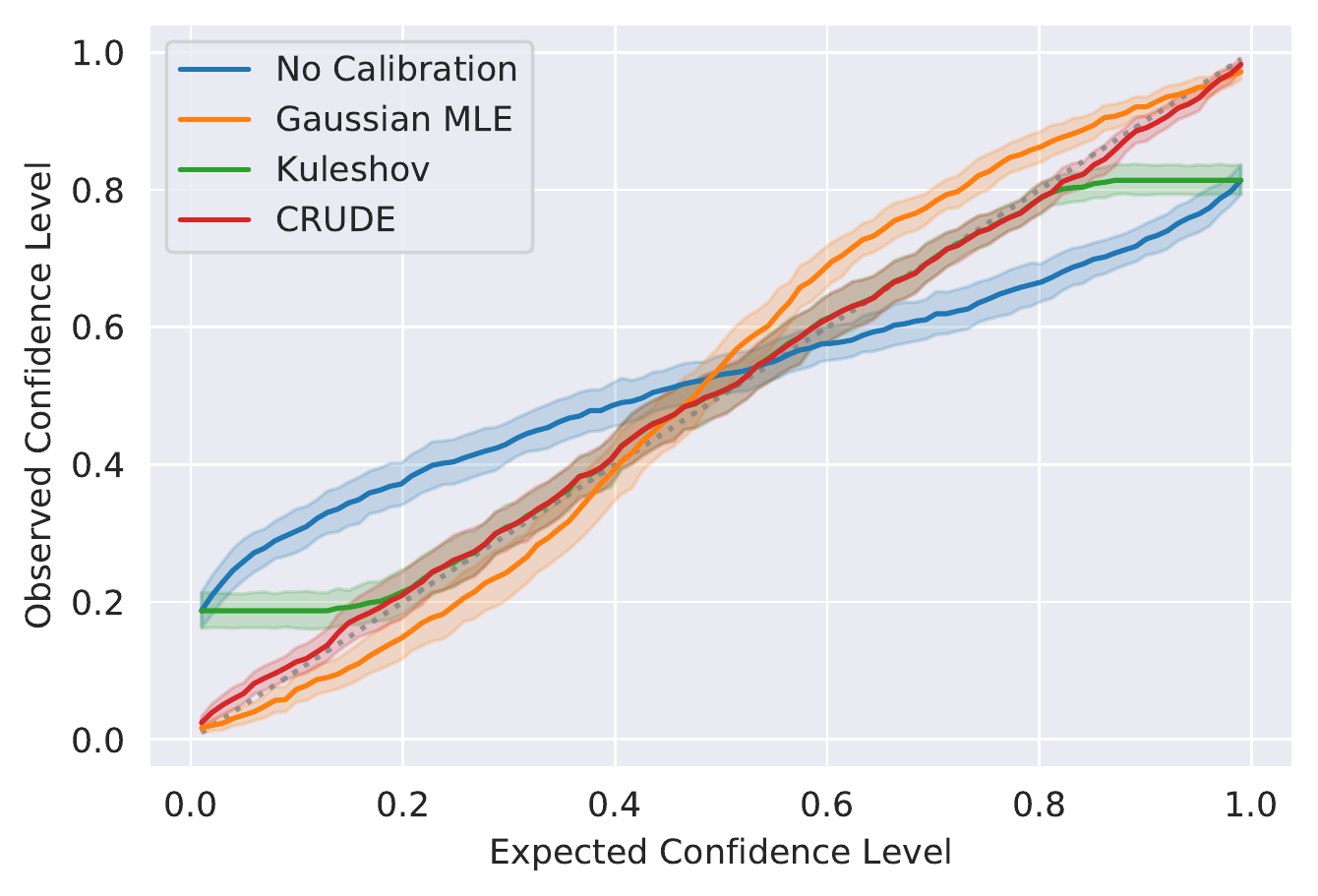} \includegraphics[width=0.39\textwidth]{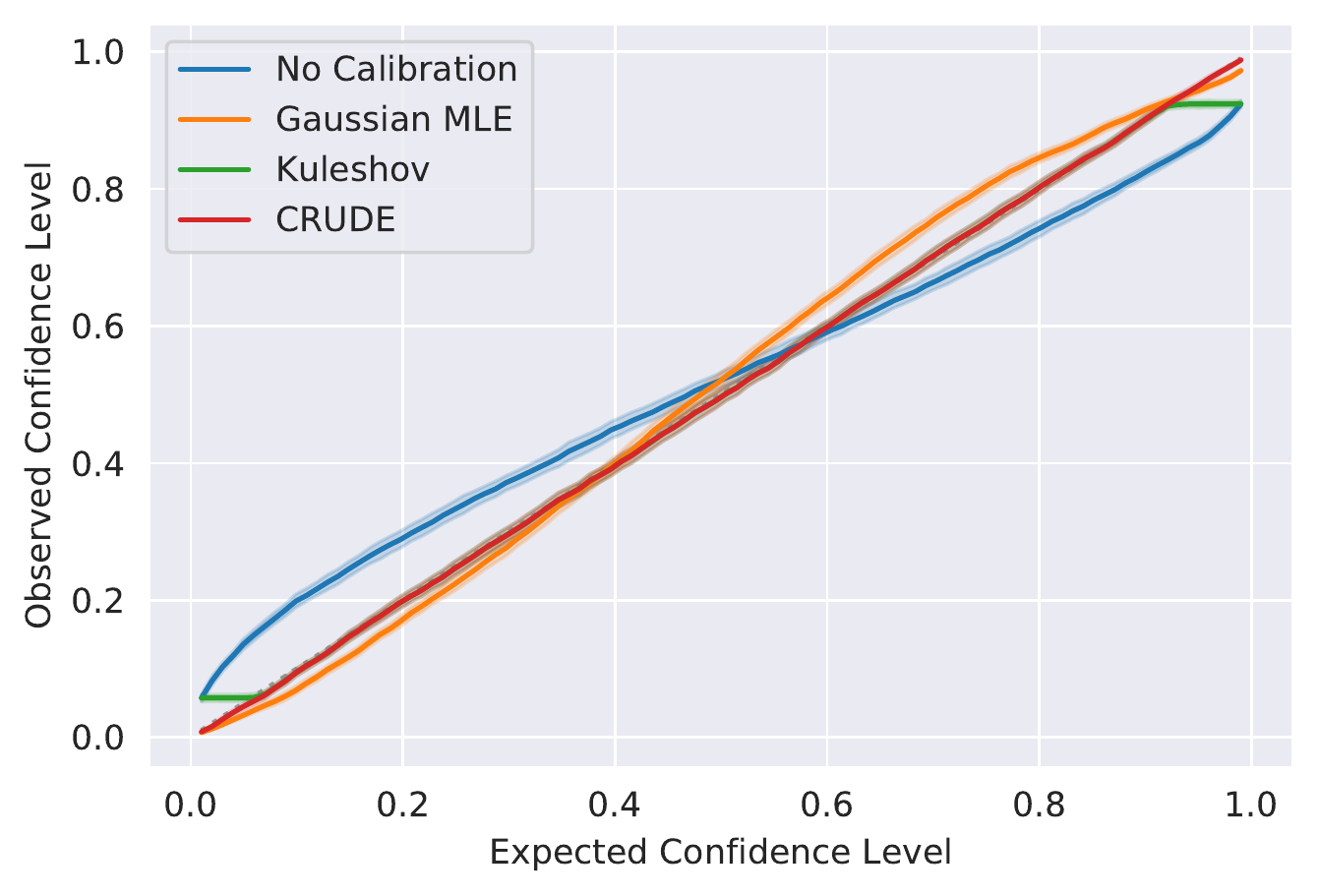}}
\centerline{\includegraphics[width=0.39\textwidth]{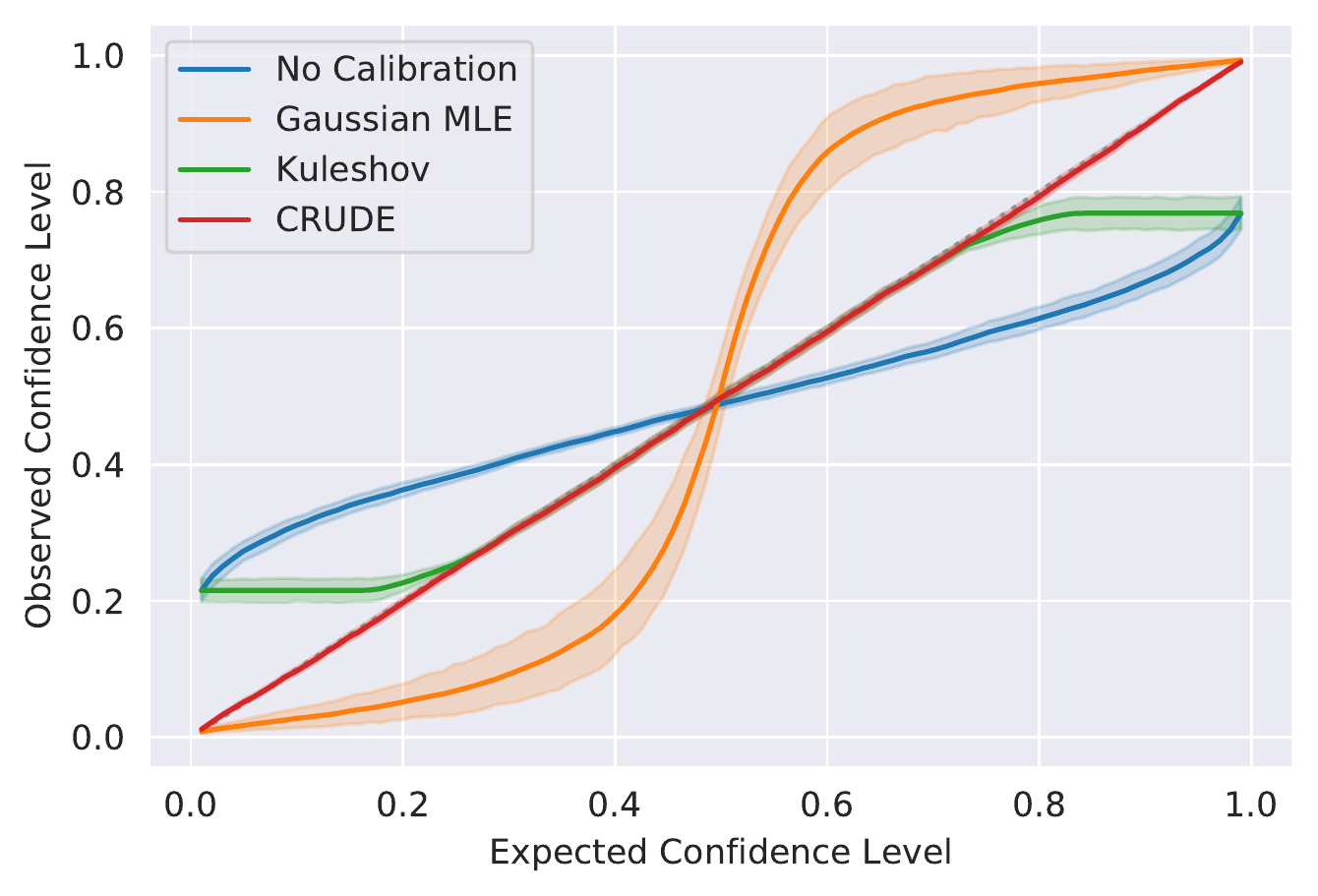} \includegraphics[width=0.39\textwidth]{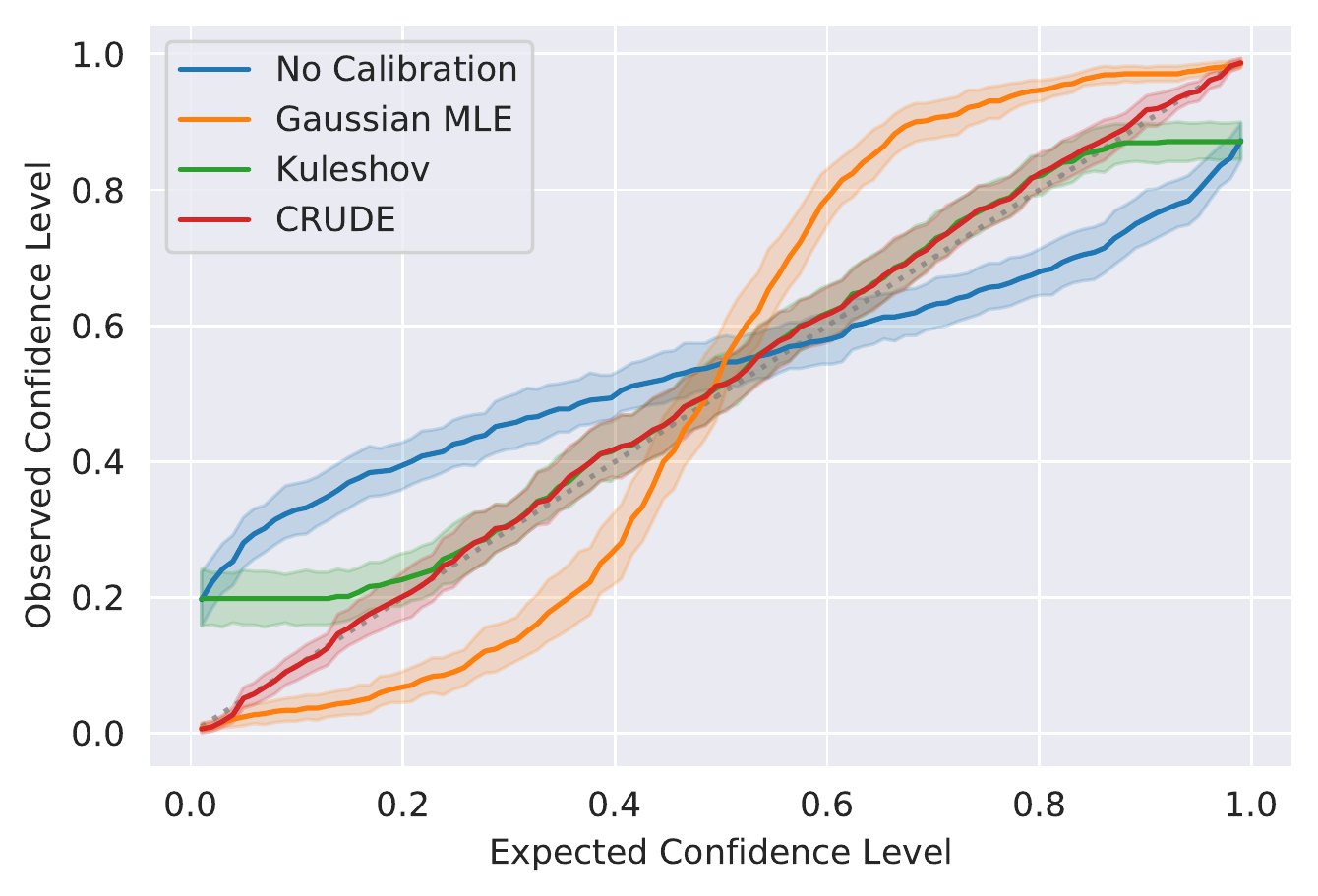}}
\caption{Various combinations of models and datasets where CRUDE performed well. In order, they are the variational neural net on the Boston Housing dataset, dropout neural net on the Parkinson's Telemonitoring dataset, Gaussian process on the Combined Cycle Power Plant datasetm and NGBoost on the Yacht Hydrodynamics dataset.}
\label{curves}
\vspace{-100px}
\end{figure}

\end{document}